\documentclass[utf8]{Manuscript}

\usepackage{url,hyperref,lineno}
\usepackage[onehalfspacing]{setspace}

\usepackage{setspace}
\usepackage{makecell}
\usepackage{hhline}
\usepackage{multirow}
\usepackage{pbox}
\usepackage{textcomp}
\usepackage[caption=false,font=footnotesize]{subfig}
\usepackage{graphicx}
\usepackage{arydshln}

\newcommand{\blue}[1]{#1}

\def\keyFont{\fontsize{8}{11}\helveticabold }
\def\firstAuthorLast{Labb\'e {et~al.}} 
\def\Authors{Mathieu Labb\'e\,$^{1,*}$, Fran{\c c}ois Michaud\,$^{1}$}
\def\Address{$^{1}$Interdisciplinary Institute of Technological Innovation (3IT),
Department of Electrical Engineering and Computer Engineering,
Universit\'{e} de Sherbrooke,
Sherbrooke, Qc, Canada  }
\def\corrAuthor{Mathieu Labb\'e}

\def\corrEmail{Mathieu.M.Labbe@USherbrooke.ca}

\def\figurename{Fig. }

\begin{document}
\onecolumn
\firstpage{1}

\title[Multi-Session for Illumination Invariant Re-Localization]{Multi-Session Visual SLAM for Illumination Invariant Re-Localization in Indoor Environments} 

\author[\firstAuthorLast ]{\Authors} 
\address{} 
\correspondance{} 

\extraAuth{}

\maketitle

\begin{abstract}

\section{}
For robots navigating using only a camera, illumination changes in indoor environments can cause re-localization failures during autonomous navigation. In this paper, we present a multi-session visual SLAM approach to create a map made of multiple variations of the same locations in different illumination conditions. The multi-session map can then be used at any hour of the day for improved re-localization capability. The approach presented is independent of the visual features used, and this is demonstrated by comparing re-localization performance between multi-session maps created using the RTAB-Map library with SURF, SIFT, BRIEF, BRISK, KAZE, DAISY and SuperPoint visual features. The approach is tested on six mapping and six localization sessions recorded at 30 minute intervals during sunset using a Google Tango phone in a real apartment.

\tiny
 \keyFont{ \section{Keywords:} Localization, Visual SLAM, Feature Matching, Mobile Robotics} 
\end{abstract}

\section{INTRODUCTION}
\label{sec:introduction}
Visual SLAM (Simultaneous Localization and Mapping) frameworks using hand-crafted visual features work relatively well in static environments, as long as there are enough discriminating visual features with moderated lighting variations. 
To be illumination invariant, a trivial solution could be to switch from vision to LiDAR (Light Detection and Ranging) sensors, but compared to cameras they are often too expensive or bulky for some applications. 
Illumination-invariant re-localization using a conventional camera is not trivial, as visual features taken during the day under natural light conditions may look quite different than those extracted during the night under artificial light conditions.
In our previous work on visual loop closure detection \citep{labbe13appearance}, we observed that when traversing multiple times the same area where atmospheric conditions are changing periodically (like day-night cycles), loop closures are more likely to be detected with locations of past mapping sessions that have similar illumination levels or atmospheric conditions. Based on that observation, in this paper, we present a multi-session approach to derive illumination invariant maps using a full visual SLAM approach. 

\blue{The idea of improving re-localization in illumination changing environments by mapping multiple times the same area \citep{dayoub2008adaptive, churchill2013experience, burki2016appearance, muhlfellner2016summary, paton2018can} is often addressed by lifelong localization systems \citep{, konolige2009towards}. A lifelong localization system would be able to adapt the map to changes in the environment to avoid degrading localization performance overtime. Doing so, there is still a risk that the robot incorrectly updates, significantly decreasing localization performance. In practice, most current navigation systems work in two phases: the SLAM phase to construct the map of the environment, then a localization-only phase when the robot navigates autonomously to accomplish its tasks without modifying the map. In this context, a human can correct gross errors in the constructed prior to launching autonomous navigation, and at some point a SLAM phase can be re-initiated to update the map overtime. Launching manually those updates could increase maintenance of the robots operating in highly dynamic environments (e.g., store, warehouse), thus a lifelong localization system would be preferred. While the approach presented in this paper shares some concepts with lifelong systems, it principally targets applications using the two-phases navigation approach for environments generally static (e.g., house, office, residence for elderly people) but having large illumination variations caused by windows or artificial lights.} 
\blue{Therefore,} the main research questions this paper \blue{focuses} are:
\begin{itemize}
  \item Which visual feature is the most robust to illumination variations in indoor environments?
  \item How many \blue{mapping} sessions are required \blue{during the SLAM phase so that} robust re-localization \blue{during the localization phase is possible} through day and night \blue{without having to update the map}?
  \item To avoid having a human teleoperate a robot at different times of the day to create the multi-session map, would it be possible to acquire the consecutive maps simply by re-localizing from the map acquired in a previous session?
\end{itemize}
By addressing these questions, the main contributions of this work are: 1) an in-depth comparison of popular visual feature approaches for illumination invariant indoor re-localization, 2) an adaptation of an Open Source SLAM framework to create multi-session maps that are robust to illumination variations, and 3) guidelines to create such multi-session maps by an autonomous robot itself.

The paper is organized as follows. Section \ref{sec:related_work} presents similar works to our multi-session map approach, which is described in Section \ref{sec:description}. Section \ref{sec:results} presents comparative results between the visual feature used and the number of sessions required to expect the best re-localization performance. Section \ref{sec:discussion} discusses limitations and possible improvements of the approach, while Section \ref{sec:conclusion} concludes this paper.

\section{RELATED WORK}
\label{sec:related_work}
The approach presented in this paper shares some similarity with the general concept of the \textit{Experience Map} \citep{churchill2013experience}. An \textit{experience} is referred to as an observation of a location at a particular time. A location can have multiple experiences describing it. New experiences of the same location are added to the experience map if re-localization fails during each traversal of the same environment. 
Re-localization of the current frame in the experience map is done concurrently against all experiences \blue{of a location}, thus requiring multi-core CPUs to do it in real-time as more and more experiences are added. \blue{To avoid examining all experiences, predicting next experiences to localize on \cite{linegar2015work, krajnik2017fremen} or selecting visually similar experiences around the current location \citep{paton2018can} can be used to test the most likely ones based on the current state of the environment.} 

To avoid using multiple experiences of the same locations, SeqSLAM \citep{milford2012seqslam, sunderhauf2013we} matches sequences of visual frames instead of trying to re-localize robustly each individual frame against the map. The approach assumes that the robot takes relatively the same route (with the same viewpoints) at the same velocity, thus seeing the same sequence of images across time. This is a fair assumption for cars (or trains), as they are constrained to follow a lane at regular velocity.
However, for indoor robots having to deal with obstacles, their path can change over time, thus not always replicating the same sequences of visual frames.

Adding more and more experiences to a map can increase its size over time, and so are computation time and memory usage. Some approaches try to limit the size of data in the map while keeping the same level of re-localization performance. In Cooc-Map \citep{johns2013feature}, local features taken at different times of the day are quantized in both the feature and image spaces, and discriminating statistics can be generated on the co-occurrences of features. This produces a more compact map instead of having multiple images representing the same location, while still having local features to recover full motion transformation. A similar approach is done in \citep{ranganathan2013towards} where a fine vocabulary method is used to cluster descriptors by tracking the corresponding 3D landmark in 2D images across multiple sequences under different illumination conditions. For feature matching with this learned vocabulary, instead of using a standard descriptor distance approach (e.g., Euclidean distance), a probability distance is evaluated to improve feature matching.
In \citep{burki2016appearance}, a selective landmark strategy is used to reduce the data bandwidth shared across a fleet of vehicles by transferring only the minimal number of landmarks from a remote multi-session map for efficient re-localization at the time the vehicle is operating. Like in our paper but for the outdoor case, they also made a specific dataset to create incrementally a multi-session map from successive trajectories taken during sunset to capture the most illumination variations. 
Similarly in \citep{muhlfellner2016summary}, a multi-session map called the Summary Map is created from merging multiple traversals of the same areas. \blue{\citep{halodova2019predictive} made an extensive comparison of map management techniques that maximize re-localization performance over time while pruning past features to limit the size of the map}. These last three papers are quite complementary to ours, where the same basic multi-session concept is used, but they are more focusing on strategies to reduce the multi-session map size than the choice of the best visual feature to use (which could also impact the map size). 
Other approaches rely on pre-processing the input images to make them illumination-invariant before feature extraction, by removing shadows \citep{mcmanus2014shady, corke2013dealing} or by trying to predict them \citep{lowry2014transforming}. This improves feature matching robustness in strong and changing shadows. \blue{In \citep{li2016hdrfusion}, auto-exposure effect is removed using a high dynamic range (HDR) map. To increase robustness against large appearance difference between seasons, \citep{neubert2013appearance} predict how the images taken during winter would look like in its map taken during summer, which improves re-localization in the same area during winter (or vice-versa)}. \blue{Most of those} approaches present results on datasets recorded outdoors with a car or a train, while in this paper we present results in an indoor setting, \blue{enhancing indoor-related works like \citep{dayoub2008adaptive, konolige2009towards, krajnik2017fremen} by explicitly addressing the robustness of re-localization in indoor illumination varying environments.}

At the local visual feature level, common hand-crafted features like SIFT \citep{Lowe04} and SURF \citep{Bay08} in outdoor experiences have been compared across multiple seasons and illumination conditions \citep{ross2013novel, valgren2007sift} to reveal some of their limitations. To overcome limitations caused by illumination variance of hand-crafted features, machine learning approaches have also been used to extract descriptors that are more illumination-invariant. \blue{In \citep{neubert2015local, krajnik2017image}, hand-crafted features have also been compared against trained descriptors, demonstrating better place recognition performance in outdoor settings.} In \citep{carlevaris2014learning}, a neural network has been trained to track interest points in time-lapse videos so that it outputs similar descriptors for the same tracked points independently of illumination. However, only descriptors are learned, and the approach still relies on hand-crafted feature detectors. More recently, SuperPoint \citep{detone2018superpoint} introduced an end-to-end local feature detection and descriptor extraction approach based on a neural network. The illumination-invariance comes from carefully making a training dataset with images showing the same visual features under large illumination variations. \blue{Other place recognition approaches using learned global descriptors exist \citep{sunderhauf2015performance, arandjelovic2016netvlad, sarlin2019coarse}, but this paper focuses more on the comparison of local (hand-crafted or learned) features that are generally used in classic visual SLAM pipelines.}

\section{MULTI-SESSION SLAM FOR ILLUMINATION INVARIANT RE-LOCALIZATION}
\label{sec:description}

\blue{The current approach is divided into two main phases: 1) the SLAM phase to construct a multi-session map containing most illumination variations of the same locations, and 2) a localization-only phase in which the robot would navigate to do its tasks using the pre-built map. In the two phases, the same re-localization approach is used, 
and in the context of SLAM the first phase is also referred to as loop closure detection. 
This section mainly describes the multi-session SLAM phase, and the differences with the localization-only phase are described at the end of the section.}

Similarly to \citep{burki2016appearance, paton2018can}, one major difference of our multi-session SLAM phase and the \textit{Experience Map} is that the interconnections of locations between the sessions are not purely topological but they also include six DoF constraints, making it possible to transform all locations in the same global coordinate frame. 
As the presented approach for the SLAM phase is targeting autonomous systems that will capture by themselves the different illumination conditions of the same environment instead of having a person teleoperating or driving the robot many times, it is preferable for the navigation system that the robot can always be localized in the same coordinate frame.
When creating the multi-session map, the mapping sessions should have enough similar illumination conditions from a previous session in order to follow correctly the original trajectory (in coordinate frame of the first session), but experience sufficient illumination variations to add new locations. As robots do not have infinite memory, the number of duplicated locations in the map should be also minimized while ideally achieving the same re-localization performance. 

\begin{figure}[!t] 
\centering 
\includegraphics[width=0.4\columnwidth]{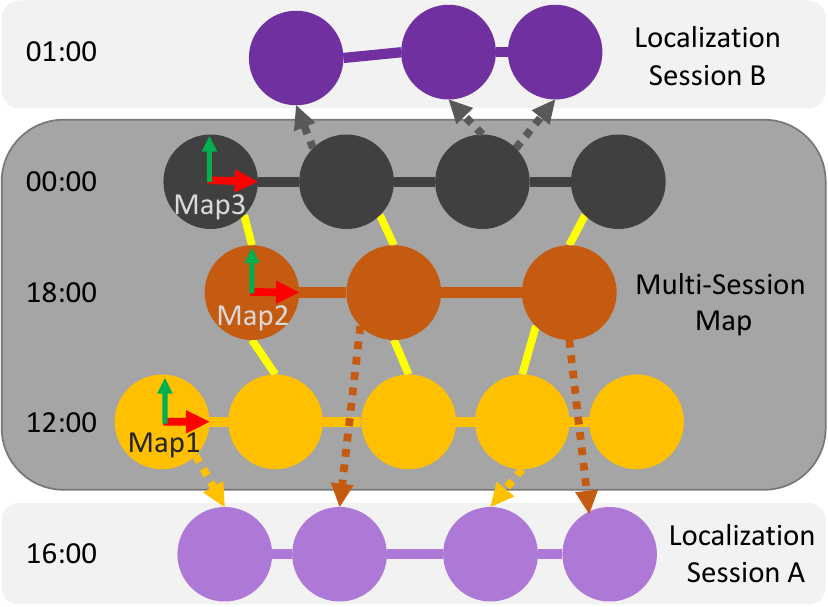} 
\caption{Structure of the multi-session map. Three sessions taken at different time have been merged together \blue{during the SLAM phase} by finding loop closures between them (yellow links). Each map has its own coordinate frame. \blue{During the localization-only phase}, Localization Session A (16:00) is re-localized in relation to both day sessions in the map \blue{(12:00 and 18:00)}, and Localization Session B (01:00) is only re-localized on the night session (00:00). } 
\label{fig:graph} 
\end{figure}

The visual feature chosen can have an impact on the final size of the multi-session map depending on how much they are illumination invariant or not. Some visual features are fast to compute and are light in memory, but a lot of them would be required to represent the different illumination states of the environment. Other visual features are more robust to illumination variation while being heavier in computation and memory, but less of them would be required to capture all variations of the environment. Therefore, the choice of visual features may impact how many sessions are required to achieve similar re-localization performance. To evaluate this, our approach is designed to be independent of the visual features used, which can be hand-crafted or neural network based, while being integrated in full SLAM conditions using for instance a library like RTAB-Map. RTAB-Map \citep{labbe18jfr} is a Graph-SLAM \citep{grisetti2010tutorial} approach that can be used with camera(s) and/or with a LiDAR. This paper focuses on the first case where only a camera is available for re-localization. The structure of the map is a pose-graph with nodes representing each image acquired at a fixed rate, and links representing the six DoF transformations between them. Figure \ref{fig:graph} presents an example of \blue{the resulting} multi-session map created \blue{during the SLAM phase} from three sessions taken at three different hours \blue{(12:00, 18:00 and 00:00)}. Two additional localization sessions are also shown, one during the day (16:00) and one at night (01:00)\blue{, which represent two examples that would be conducted during the localization-only phase}. The dotted links represent to which nodes in the graph the corresponding frame has been re-localized on. 
The goal is to have new frames re-localizing on nodes taken at a similar time, and if localization time falls between two mapping times, re-localization could jump between two sessions or more inside the multi-session map. 
Inside the multi-session map, each individual map are transformed in the same global coordinate frame (Map1 in this example) so that when the robot re-localizes on a node of a different session, it does not jump between different coordinate frames.

\begin{figure}[!t] 
\centering 
\includegraphics[width=\textwidth]{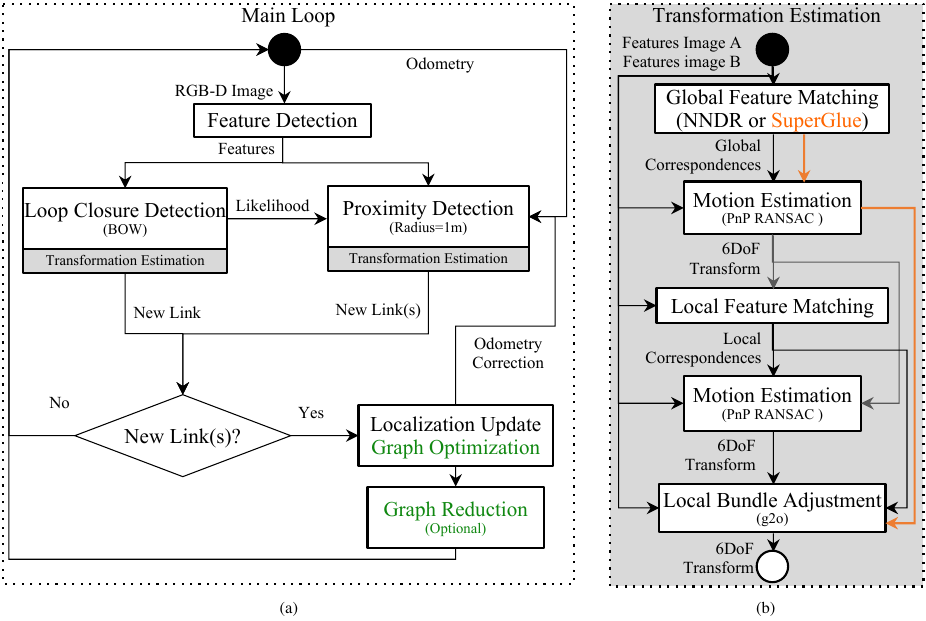} 
\caption{Main loop of the SLAM approach used. In green are steps only done during multi-session SLAM phase when combining the mapping sessions. During localization phase, only Localization Update is done if \blue{new link} (re-localization) has been \blue{added} (the graph is not modified, \blue{only odometry correction is applied}).} 
\label{fig:main_loop} 
\end{figure}


Figure \ref{fig:main_loop} presents the main loop of the the SLAM algorithm \blue{used during SLAM phase}, which can be done online or offline. After a new frame and its pose are received, visual features are extracted from the RGB image with their 3D positions estimated using the depth image and camera calibration. Visual features can be any of the ones implemented in OpenCV \citep{bradski2008learning}, which are SURF \citep{Bay08}, SIFT \citep{Lowe04}, BRIEF \citep{calonder2010brief}, BRISK \citep{leutenegger2011brisk}, KAZE \citep{alcantarilla2012kaze} and DAISY \citep{tola2009daisy}. The SuperPoint \citep{detone2018superpoint} neural network based feature has also been integrated for comparison. 

Two methods are used to find loop closures: a global one called Loop Closure Detection (LCD), and a local one called Proximity Detection (PD). \blue{LCD is not limited to only nodes of the current mapping session, but it also includes all nodes from all past sessions when updating its loop closure hypotheses. This makes the approach able to seamlessly find constraints between sessions that are used to merge multiple sessions together during the Graph Optimization step.} The bag-of-words (BOW) approach \blue{\citep{sivic2003video}} is used to evaluate loop closure hypotheses over all previous images from all sessions, independently of the odometry estimate. The BOW vocabulary \blue{used in this paper} is incremental based on FLANN's KD-Trees \citep{muja_flann_2009}, and quantization of features to visual words is done using the Nearest Neighbor Distance Ratio (NNDR) approach \citep{Lowe04}. \blue{After quantization of the features of the current frame into the vocabulary, BOW uses an inverted index voting-scheme to retrieve past images with the same visual words, to significantly reduce the likelihood estimation time with all previous images. The likelihood is then fed to a Bayes filter to estimate loop closure hypotheses \citep{labbe13appearance}. The Bayes filter helps filter spurious wrong likelihood (because of noise), so that a node in the map should score high in likelihood on many consecutive frames for its hypothesis to grow.} When a loop closure hypothesis reaches a pre-defined threshold, a loop closure is detected. In contrast to LCD, PD looks for nodes around the current position of the robot for possible loop closures based on the current odometry estimate. Nodes of the map's graph inside a fixed radius of the current position are then selected as candidates for proximity detection. In our previous work \citep{labbe2017}, proximity detection was introduced primarily in RTAB-Map for LiDAR rotational invariant re-localization. A slight modification is made for this work to use it with a camera. Previously, the closest nodes in a fixed radius around the current position of the robot were sorted by distance, then PD was used against the top three closest nodes, adding the same number of constraints to the graph if all of them are accepted. However, in a visual multi-session map, the closest nodes may not have images with the most similar illumination conditions than the current one. \blue{Similar to BOW selection in \citep{paton2018can},} by using the likelihood computed during LCD, nodes inside the proximity radius are sorted from the most to less visually similar (in terms of BOW's inverted index score). Visual PD is then done using the three most similar images around the current position in a fixed radius. If PD fails because the robot's odometry drifted too much since the last re-localization, LCD is still done in parallel to re-localize the robot when it is lost. 

For both loop closures and proximity detections, six DoF transformations are computed following the steps of Transformation Estimation (TE) of \figurename \ref{fig:main_loop}. A global feature matching is done using \blue{a nearest neighbor (NN) approach with feature descriptors} between the corresponding frames. With the feature correspondences, a first transformation between frames is computed using the Perspective-n-Point (PnP RANSAC) approach \citep{bradski2008learning}. Using that previous transformation as a motion estimate, 3D features from the first frame are then projected into the second frame for local feature matching using a fixed size window. This second step generates better matches to compute a more accurate transformation using PnP. \blue{Depicted by oranges arrows in \figurename \ref{fig:main_loop}, if the visual feature type used is SuperPoint, the SuperGlue approach \citep{sarlin20superglue} can be optionally used for global feature matching. SuperGlue uses a neural network trained to find correspondences between SuperPoint features, generating more correspondences than classic NNDR approach. In that case, the second local feature matching step along with the second motion estimation step are skipped. For both approaches,} the resulting transform is further refined using a local bundle adjustment approach \citep{kummerle11g2o}.

When loop closures are detected, the pose-graph is optimized using GTSAM \citep{dellaert2012factor} with the new constraints, implicitly transforming all sessions into the same coordinate frame as long as there is at least one loop closure between the sessions. This means that when a loop closure happens for the first time with an older session, the whole current map is automatically transformed in the coordinate frame of the oldest map. This may cause large re-localization jumps when these events happen. However, once the maps are merged, the next re-localization jumps should be proportional to odometry drift and how long the robot has not been re-localized. Finally, a Graph Reduction (GR) approach can be used to reduce the size of the map when loop closures have been previously added to graph, thus reducing memory usage of the algorithm. This process is explained in details in \citep{labbe2017} and is similar to approach in \citep{churchill2013experience} where if re-localization is successful, no new experiences are added. In summary, a node having a loop closure with an older node can be removed by merging the loop closure links to its neighbor nodes, thus keeping the graph at the same size when new data is acquired as long as there are loop closures. The graph will increase in size only when a loop closure has not been detected (e.g., the location has changed too much or a new area is visited).

After the multi-session map is created, the localization-only phase is done following the same main loop than \figurename \ref{fig:main_loop}, but without the green steps (the pose-graph is not modified). Another difference is that to limit processing time, when estimating transformations of the top three identified nodes from LCD and PD, as soon as a first transformation is accepted the others are not tested. \blue{In the next section, both LCD and PD are referred to as re-localization during the localization-only phase.}

\section{RESULTS}
\label{sec:results}

\begin{figure*}[!t] 
\centering 
\includegraphics[width=\textwidth]{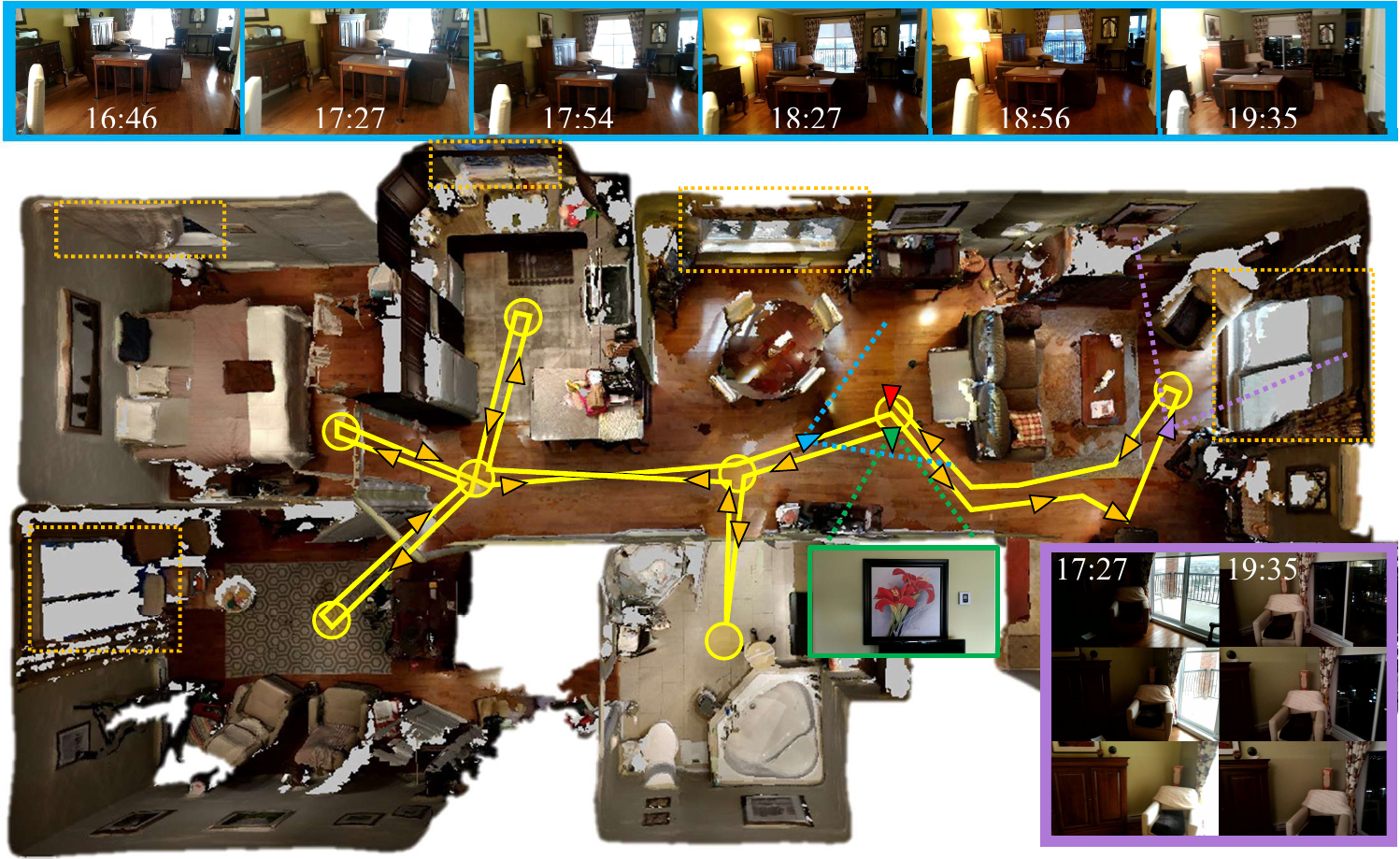} 
\caption{Top view of the testing environment with the followed trajectory in yellow. Start and end positions correspond to green and red triangles respectively, which are both oriented toward same picture on the wall shown in the green frame. Circles represent waypoints where the camera rotated in place. Windows are located in the dotted orange rectangles. The top blue boxes are pictures taken from a similar point of view (located by the blue triangle) during the six mapping sessions. The purple box shows three consecutive frames from two different sessions taken at the same position (purple triangle), illustrating the effect of auto-exposure.} 
\label{fig:overview} 
\end{figure*}

To address the three research questions presented in Section \ref{sec:introduction}, a dataset has been recorded before and after sunset to capture the full spectrum of illumination variations between the day and night. Figure \ref{fig:overview} illustrates how the dataset has been acquired in a home in Sherbrooke, Quebec in March 2019. An ASUS Zenfone AR phone (with Google Tango technology) running the RTAB-Map Tango App has been used to record data for each session following the same yellow trajectory, similarly to what a robot would do patrolling the environment. The poses are estimated using Google Tango's visual inertial odometry approach, with RGB and registered depth images recorded at 1 Hz.
\blue{To be able to combine offline the maps into multi-session maps, as described in Section \ref{sec:multisession}}, the trajectory started and finished in front of a highly visual descriptive location (i.e., first and last positions shown as green and red arrows, respectively) to make sure that each consecutive mapping session is able to re-localize on start from the previous session. Note that this assumption could be also valid for a robot by placing a highly discriminating sign visible at its docking station. This ensures that all maps are transformed in the same coordinate frame of the first map after graph optimization. Between 16:45 (daylight) and 19:45 (nighttime), two mapping sessions were recorded back-to-back to get a mapping session and a localization session taken roughly at the same time. The time delay between each mapping session is around 30 minutes. Overall, the resulting dataset has six mapping sessions (Numerical Index-Time: 1-16:46, 2-17:27, 3-17:54, 4-18:27, 5-18:56, 6-19:35) and six localization sessions (Alphabetical Index-Time: A-16:51, B-17:31, C-17:58, D-18:30, E-18:59, F-19:42). 

The top blue boxes of \figurename \ref{fig:overview} show images of the same location taken during each mapping session. To evaluate the influence of natural light coming from the windows during the day, all lights in the apartment were on during all sessions except for one in the living room that we turned on when the room was getting darker (see top images at 17:54 and 18:27). Beside natural illumination changing over the sessions, the RGB camera had auto-exposure and auto-white balance enabled (which could not be disabled by Google Tango API on that phone), causing additional illumination changes depending on where the camera was pointing, as shown in the purple box of \figurename \ref{fig:overview}. The left sequence (17:27) illustrates what happened when greater lighting comes from outside, with auto-exposure making the inside very dark when the camera passed by the window. In comparison, doing so at night (shown in the right sequence 19:35) did not result in any changes. Therefore, for this dataset, most illumination changes are coming either from natural lighting or auto-exposure variations. 

For the implementation, OpenCV 4.2.0 and RTAB-Map 0.20.15 have been used. Table \ref{table:parameters} presents RTAB-Map's parameters used. Note that "Kp/MaxFeatures" parameter means that only 400 features of the 1000 extracted from each frame (``Vis/MaxFeatures'') with highest response are quantized to BOW vocabulary, to limit vocabulary size over time. Experimentally, we found that ``Vis/CorNNDR=0.6'' works better when features are more discriminative (i.e., have float descriptors) and set to $0.8$ for binary features. The feature detector value can be SURF (SU), SIFT (SI), BRIEF (BF), BRISK (BK), KAZE (KA), DAISY (DY) and SuperPoint (SP). \blue{The SuperPoint variant with SuperGlue feature matching is named SG. Note that results using other binary features available in OpenCV like ORB \cite{rublee2011orb} and FREAK \cite{alahi2012freak} are very similar to BRIEF in terms of processing time, memory and re-localization performance, and therefore only BRIEF results are presented in this paper.}

\begin{table}[!t]
\caption{RTAB-Map's parameters.}
\label{table:parameters}
\centering
\setlength\tabcolsep{1.5pt} 
\begin{tabular}{|l|l|r|}
\hline
{\textbf{Name}} & {\textbf{Description}} & {\textbf{Value}} \\
\hline
Kp/DetectorStrategy & Feature detector & \textit{Variable} \\
Kp/MaxFeatures & Maximum visual words per frame & 400 \\
Vis/CorGuessWinSize & Local feature matching window size & 40 pix \\
Vis/CorNNDR & NNDR for binary features & 0.8 \\
Vis/CorNNDR & NNDR for float features & 0.6 \\
Vis/CorNNType & Feature matching approach & 1 (NN) or 6 (SG)\\
Vis/FeatureType & Feature detector & \textit{Variable}  \\
Vis/MaxFeatures & Maximum visual features per frame & 1000 \\
Vis/MinInliers & Minimum PnP inliers & 20 \\
Reg/RepeatOnce & Second local feature matching & true (false with SG)\\
RGBD/LocalRadius & Proximity detection local radius & 1 m \\
RGBD/ProximityMaxPaths & Maximum nodes tested by proximity detection & 3 \\
Rtabmap/LoopThr & Loop closure detection threshold & 0.11 \\
\hline
\end{tabular}
\end{table}

\blue{To evaluate re-localization performance, the metric used is the percentage of frames that are re-localized during the localization phase, i.e., the number of frames correctly re-localized on the total number of frames in a localization session. For example, if the localization session has 300 frames and only 200 frames are re-localized, localization performance is 66\%. 
A correct re-localization means that the localized frame represents the same real location than the corresponding frame in the map. In all our experiments below, no wrong re-localized frames were accepted by the algorithm, because similarity was insufficient to trigger a re-localization (LCD hypotheses $<$ Rtabmap/LoopThr) or that TE rejected them because of lack of visual inliers ($<$ Vis/MinInliers).}

\subsection{Single Session Re-Localization}

\begin{figure*}[!t] 
\centering 
\includegraphics[width=\textwidth]{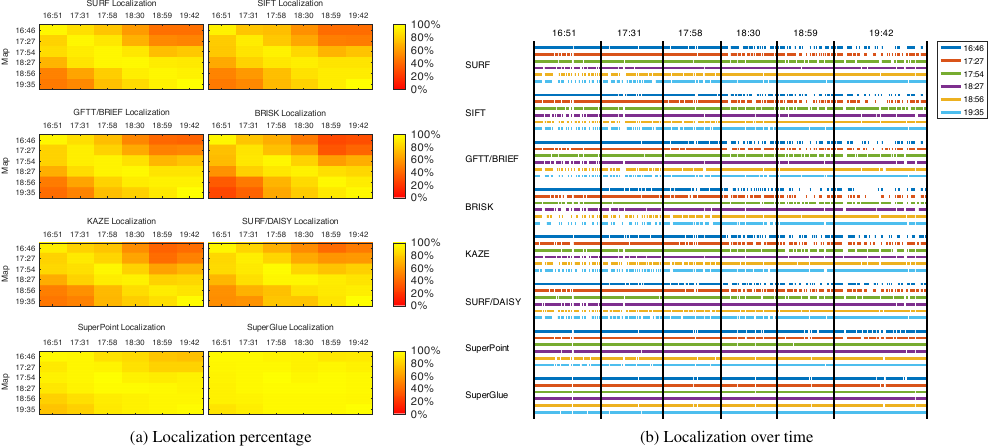} 
\caption{Re-localized frames over time of the A to F localization sessions ($x$-axis) over the 1 to 6 single-session maps ($y$-axis) in relation to the visual features used: a) re-localization percentage; b) re-localization over time.} 
\label{fig:single_results} 
\end{figure*}

The first experiment done with this dataset examines re-localization performance of different visual features for a single mapping session, thus establishing our baseline performance. 
Figure \ref{fig:single_results}a shows the percentage of frames re-localized of the localization sessions (A to F) over the mapping sessions (1 to 6) independently, for each visual features listed in Section \ref{sec:description}. Figure \ref{fig:single_results}b shows more precisely when every frames have been re-localized on each mapping session, thus visualizing the distribution of the re-localization.  As expected by looking at the diagonals, re-localization performance is best (and with less gaps) when re-localization is done using a map taken at the same time of the day (i.e., with very similar illumination condition). In contrast, re-localization performance is worst when re-localizing at night using a map taken the day, and vice-versa. 
SuperPoint \blue{(with or without SuperGlue)} is the most robust descriptor to large illumination variations, while binary descriptors like BRIEF and BRISK are the most sensitive.

\subsection{Multi-Session Re-Localization}
\label{sec:multisession}

\begin{figure*}[!t] 
\centering 
\includegraphics[width=\textwidth]{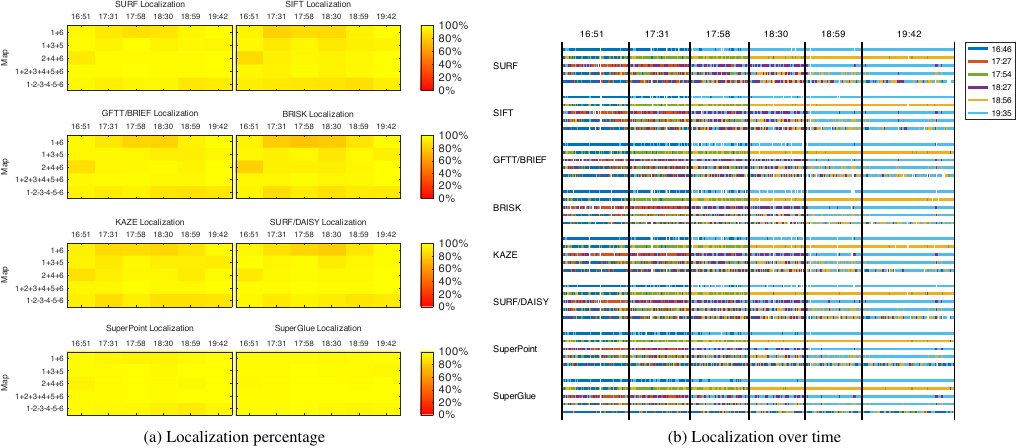} 
\caption{Re-localized frames over time of the A to F localization sessions ($x$-axis) over the five multi-session maps 1+6, 1+3+5, 2+4+6, 1+2+3+4+5+6 and 1-2-3-4-5-6 ($y$-axis, ordered from top to bottom), in relation to the visual features used: a) re-localization percentage; b) re-localization over time.} 
\label{fig:merged_results} 
\end{figure*}

The second experiment evaluates re-localization performance using multi-session maps created using maps generated at different times from the six mapping sessions. \blue{To create different combinations of multi-session maps from the six individual map sessions recorded, the selected individual maps are replayed back to back offline as input streams to a new SLAM process. This new SLAM process can detect the transition between input maps to internally create a new session. Because all mapping sessions started in front of the same highly visual descriptive location, LCD can detect a loop closure with previous session in order to merge, through graph optimization, the internal sessions in the same global graph. As more input data are streamed to the new SLAM process, more loop closures are detected between and inside sessions.}
Different combinations \blue{of multi-session maps} are tested: 1+6 combines the two mapping sessions with the highest illumination variations (time 16:46 with time 19:35); 1+3+5 and 2+4+6 are multi-session maps generated by assuming that mapping would occur every hour, and 1+2+3+4+5+6 is the combination of all maps taken at 30 minutes interval. Those \blue{multi-session} maps have been merged without graph reduction, thus keeping all nodes of all sessions. For comparison, the multi-session map 1-2-3-4-5-6 represents assembled maps with graph reduction enabled. Figure \ref{fig:merged_results} illustrates, for each multi-session map, the resulting re-localization performance and re-localized frames over time. Except for SuperPoint \blue{(with and without SuperGlue)} which shows similar high performance for all multi-session maps, merging more sessions with different illumination conditions increases re-localization performance for all visual features, with best performance using the 1+2+3+4+5+6 multi-session map and with 1-2-3-4-5-6 not far behind. Note that for those multi-session results, \figurename \ref{fig:merged_results}b also shows by color to which map in the tested multi-session map a frame has been re-localized. For example, for 2+4+6 and 1+2+3+4+5+6 maps (third and fourth lines), most re-localizations after 19:42 have been on frames added from Map 6 (19:35). For the reduced map 1-2-3-4-5-6 (last line), re-localizations are more distributed against all \blue{mapping} sessions for every localization sessions.

\begin{table*}[!t]
\caption{Cumulative re-localization performance (\%) and average re-localization jumps (mm) of the six localization sessions on each map for each visual feature used.}
\label{cumulative_loc}
\centering
\setlength\tabcolsep{1pt} 
\begin{tabular}{lcccccccc|ccccccccc|ccccccccc}
& \multicolumn{8}{c}{Re-Localization (\%)} && \multicolumn{8}{c}{Visual Inliers (\%)} &&  \multicolumn{8}{c}{Jumps (mm)} \\
Map         & SU & SI & BF & BK & KA & DY & SP & SG        && SU & SI & BF & BK & KA & DY & SP & SG  && SU & SI & BF & BK & KA & DY & SP & \blue{SG}\\
\hline
1           & 68 & 65 & 65 & 58 & 65 & 66 & 87 & 94        && 11 & 10 & 15 & 11 & 12 & 11 & 19 & 23  && 58 & 56 & 53 & 60 & 54& 45 & 39 & 41\\
2           & 74 & 69 & 71 & 62 & 69 & 74 & 89 & 96        && 11 & 10 & 14 & 10 & 12 & 10 & 19 & 23  && 41 & 38 & 41 & 46 & 42 & 37 & 36 & 36\\
3           & 86 & 82 & 84 & 78 & 81 & 86 & 95 & 98        && 13 & 11 & 16 & 11 & 14 & 13 & 23 & 27  && 43 & 43 & 42 & 48 & 50 & 41 & 33 & 32\\
4           & 89 & 85 & 87 & 82 & 84 & 88 & 96 & 98        && 13 & 12 & 16 & 11 & 15 & 13 & 23 & 29  && 38 & 48 & 42 & 44 & 41 & 36 & 31 & 31\\
5           & 80 & 76 & 79 & 71 & 7  & 79 & 91 & 96        && 13 & 12 & 16 & 12 & 15 & 13 & 22 & 27  && 39 & 45 & 46 & 49 & 46 & 39 & 34 & 36\\
6           & 76 & 75 & 74 & 67 & 72 & 76 & 90 & 96        && 13 & 12 & 17 & 12 & 15 & 13 & 22 & 26  && 41 & 46 & 44 & 45 & 43 & 37 & 36 & 36\\
1+6         & 94 & 91 & 92 & 87 & 90 & 90 & 97 & \textbf{99}          && 14 & 13 & 17 & 12 & 17 & 14 & 25 & 30  && 38 & 37 & 38 & 46 & 36 & 37 & 29 & 31\\
1+3+5       & 97 & 95 & 96 & 95 & 94 & 97 & \textbf{99} & \textbf{99} && 16 & 14 & 18 & 13 & 18 & 16 & 27 & 32  && 31 & 57 & 37 & 44 & 42 & 32 & 29 & 28\\
2+4+6       & 98 & 95 & 96 & 93 & 95 & 97 & 98 & \textbf{99}          && 16 & 14 & 19 & 13 & 18 & 16 & 27 & 32  && 32 & 37 & 38 & 38 & 38 & 31 & 27 & 27\\

1+2+3+4+5+6 & \textbf{99} & \textbf{97} & \textbf{98} & \textbf{98} & \textbf{98} & \textbf{99}  & \textbf{99} & \textbf{99}&& 
\textbf{18} & \textbf{16} & \textbf{20} & \textbf{14} & \textbf{20}  & \textbf{18} & \textbf{29} & \textbf{34} && 
\textbf{26} & \textbf{29} & \textbf{34} & 37 & 32  & \textbf{28} & \textbf{25} & \textbf{24} \\

1-2-3-4-5-6 & 94 & 91 & 92 & 91 & 90 & 93 & 96 & 98        && 12 & 12 & 16 & 11 & 14 & 12 & 21 & 23   && 43 & 45 & 51 & 52 & 48 & 43 & 40 & 44\\

1\textbar2\textbar3\textbar4\textbar5\textbar6 & 97 & 95 & 96 & 95 & 95 & 96 & 98 & 98 && 
17 & 14 & 19 & \textbf{14} & 18 & 16 & 27 & 31 && 
27 & 32 & 32 & \textbf{35} & \textbf{31} & \textbf{28} & 26 & 27\\
\hline
\end{tabular}
\end{table*}

Table \ref{cumulative_loc} presents cumulative re-localization performance over single session and multi-session maps for each visual feature. 
As expected, multi-session maps \blue{improve} re-localization performance.
The map 1\textbar2\textbar3\textbar4\textbar5\textbar6 corresponds to the case when only the session map taken at the closest time of each localization session is used (e.g., the cumulative performance of the diagonal results in \figurename \ref{fig:single_results}a).
While this seems to result in similar performance compared to the multi-session cases, this could be difficult to implement robustly over multiple seasons (where general illumination variations would not always happen at the same time) and during weather changes (cloudy, sunny or rainy days would result in changes in illumination conditions for the same time of day). 
Another challenge is how to make sure that maps remain correctly aligned together in the same coordinate frame \blue{of the original map and also} during the whole trajectory. \blue{A global localization drift could then happen over time, which is referred to as the 'photocopy' of a 'photocopy' effect \citep{halodova2019predictive}.}
In contrast, with the multi-session approach, the selection of which mapping session to use is done implicitly by selecting the best candidates of loop closure detection across all sessions. There is therefore no need to have a priori knowledge of the illumination conditions before doing re-localization. \blue{All sessions are also correctly aligned with regard to the origin of the original map.}

Even if we did not have access to ground truth data recorded with an external global localization system, the odometry from Google Tango for this kind of trajectory and environment does not drift very much. Thus, evaluating re-localization jumps caused by odometry correction can provide an estimate of re-localization accuracy. The last columns of Table \ref{cumulative_loc} present the average distance of the jumps occurring during localization. The maps 1+2+3+4+5+6 and 1\textbar2\textbar3\textbar4\textbar5\textbar6 produce the smallest jumps, which can be explained by three factors: 1) the high number of visual inliers (middle columns) when computing the transformation between two frames; 2) smaller gaps between re-localized frames (that would increase odometry drift); and 3) the presence of more frames with the same illumination level in the map. With these two maps, re-localization frames can be matched with map frames taken roughly at the same time, thus giving more and better inliers.

%
\begin{table}[!t]
\caption{Graph size and RAM usage (MB) computed using Valgrind's Massif tool of each map for each visual feature used. }
\label{ram_usage}
\centering
\begin{tabular}{ll|rrrrrrrr}
Map & Nodes & SU & SI & BF & BK & KA & DY & SP & \blue{SG} \\
\hline
1 & 202       & 104 & 167  & 64  & 80  & 91  & 233 & 205 & 205 \\
2 & 201       & 101 & 162  & 64  & 79  & 83  & 227 & 189 & 189 \\
3 & 204       & 103 &   165  &   64 &    80  &   87   &  231  &  195 & 195 \\
4 & 191       & 94  &  151   &  60  &   74  &   80   &  211  &  183 & 183  \\
5 & 175       & 85  &  134   &  55  &   67  &   71    & 189  &  161 & 161 \\
6 & 234       & 111  &  176  &   71  &   86  &   96   &  251 &   200 & 200 \\
\hline
1+6 & 436     & 215  &  343  &  134  &  166  &  186  &  483 &   404 & 404 \\
1+3+5 & 581   & 288  &  457  &  179  &  223  &  244  &  644 &   542 & 542  \\ 
2+4+6 & 626   & 303  &  481  &  191  &  236  &  253  &  681  &  559 & 559 \\ 
1+2+3+4+5+6 & 1207   & 583 &   924   & 367 &   454  &  489  &  1303 &  1077 & 1077 \\ 
1-2-3-4-5-6 & Table \ref{reduced_size}   & 176  &  308   & \textbf{128}   & 169 &   162   &  404 &   252 & 184 \\
\hline
Constant Overhead &   & 90 & 90 & 90 & 225 & 90 & 155 & 1445 & 1445 \\
\hline
\end{tabular}
\end{table}

\begin{table}[!t]
\caption{Graph size for the 1-2-3-4-5-6 multi-session map for each visual feature used and the percentage of nodes removed in comparison to 1+2+3+4+5+6 map. }
\label{reduced_size}
\centering
\begin{tabular}{l|cccccccc}
& SU & SI & BF & BK & KA & DY & SP & \blue{SG} \\
\hline
Nodes & 349 & 395 & 401 & 426 & 395 & 362 & 272 & \textbf{198} \\
Reduction & 71\% & 67\% & 67\% & 65\% & 67\% & 70\% & 77\% & \textbf{84\%} \\ 
\hline
\end{tabular}
\end{table}

\begin{table}[!t]
\caption{Average re-localization time and features per frame for each visual feature used, along with descriptor dimension and number of bytes per element in the descriptor.}
\label{average_time}
\centering
\setlength\tabcolsep{5pt} 
\begin{tabular}{llrrrrrrrr}
& & SU & SI & BF & BK & KA & DY & SP & \blue{SG}\\
\hline
Feature Detection (ms) & & 39 &  152 &   15 &  332  &  397 & 89 &  85 & 85 \\ \hdashline[0.5pt/5pt]
\multirow{3}{*}{Loop Closure Detection (ms)} & with single-session maps & 7 & 8 & 7 & 8 & 7 & 8 & 8 & 8\\
                     & with 1+2+3+4+5+6 map & 11 & 11 & 11 & 12 & 10 & 11 & 11 & 11\\
                     & with 1-2-3-4-5-6 map & 8 & 8 & 8 & 9 & 7 & 9 & 8 & 8\\ \hdashline[0.5pt/5pt]
Transformation Est. (ms) & & 35 & 30 & 37 & 36 & 26 & 37 & 24 & 64\\
\hline
Features/Frame  & & 847 &  770 &   889 &  939 &  640 & 847&  472 & 472 \\\hdashline[0.5pt/5pt]
\multirow{3}{*}{Vocabulary Size (x$10^3$)} & with single-session maps & 42 & 46 & 59 & 58 & 40 & 39 & 35 & 35\\
                            & with 1+2+3+4+5+6 map & 231 & 248 & 331 & 327 & 214  & 209 & 181 & 181\\
                            & with 1-2-3-4-5-6 map & 78 & 92 & 119 & 127 & 78 & 75 & 49 & 35\\ \hdashline[0.5pt/5pt]
Descriptor Dimension & & 64 &  128 &   32 &  64 &  64  & 200 &  256 & 256 \\
Descriptor Bytes/Elem & & 4 &  4 &   1 &  1 &  4  & 4 &  4 & 4 \\
\hline
\end{tabular}
\end{table}

Regarding computation resources, the multi-session approach requires more memory usage, as the map is at most six times larger in our experiment than a single session of the same environment if graph reduction is not applied. 
Table \ref{ram_usage} presents the memory usage (RAM) required for re-localization using the different maps configurations, along with the constant RAM overhead (e.g., loading libraries and feature detector initialization) shown separately at the bottom. 
Graph reduction with SuperPoint is higher than other features (see Table \ref{reduced_size}), which can be explained by being the most illumination invariant feature, causing more nodes to be reduced. \blue{With SuperGlue, as more feature correspondences can be found between frames, thus accepting more re-localizations, the reduction is higher and the final map size is even smaller than most individual map sessions. From the 198 nodes remaining in the final reduced map, 144 nodes are coming from Map 1, 9 from Map 2, 13 from Map 3, 8 from Map 4, 12 from Map 5 and 11 from Map 6.} 
Table \ref{average_time} presents the average re-localization time (on an Intel Core i7-9750H CPU and a GeForce GTX 1650 GPU for SuperPoint \blue{and SuperGlue}). Feature detection time depends only on the feature type, and with all maps, this is what takes the most processing time per frame. TE time is also independent of the map size, but dependent on the number of features extracted per frame. Using BOW's inverted index search, loop closure detection computation does not require significantly more time to process for multi-session maps (at most +4 ms for graph and vocabulary six times larger) than for single-session maps. However, the multi-session maps require more memory, which could be a problem on small robots with limited RAM. With graph reduction, memory usage can be reduced to a level between single-session and two-session maps. Comparing the visual features used, BRIEF requires the least processing time and memory. Even if it generates the most features per frame and a larger vocabulary, its descriptor is so small that less RAM is used. TE time is also the lowest with SuperPoint, as less features are extracted per frame. However, it requires significantly more memory (even more than multi-session maps of other features without graph reduction) because of its high dimensional descriptor and a large NVidia's CUDA librarie overhead in RAM. \blue{SuperGlue adds a 40 ms overhead on TE when used.}

\begin{figure*}[!t] 
\centering 
\includegraphics[width=\textwidth]{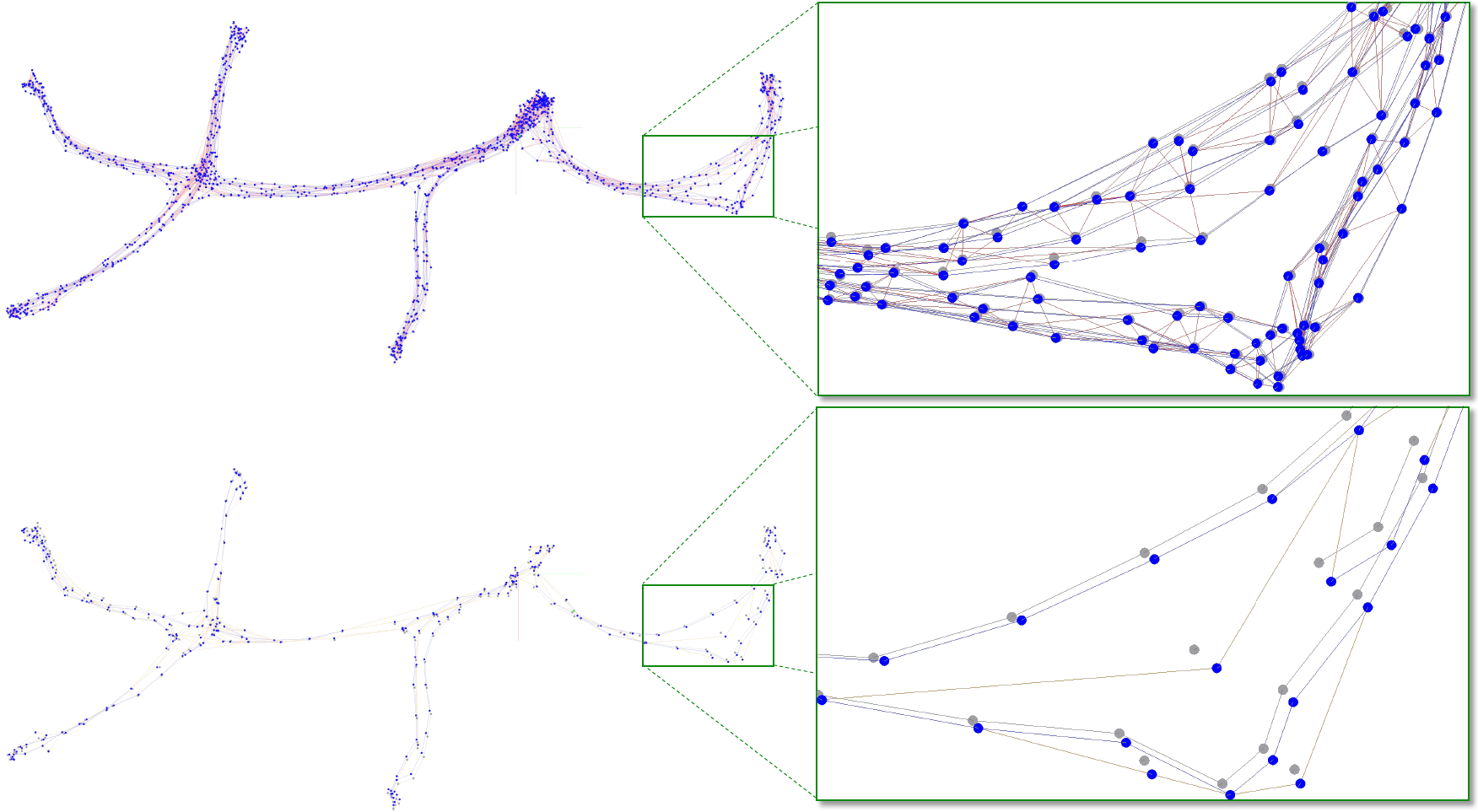} 
\caption{Comparison of the multi-session maps 1+2+3+4+5+6 (top) and 1-2-3-4-5-6 (down) with SuperPoint feature. On the right are the zoomed parts of the corresponding rectangles on the left.  Gray nodes correspond to what is considered to be the ground truth. Loop closure and odometry links are shown in blue and red, respectively. Orange links are created when reducing the graph (loop closure links propagated to neighbor nodes when a node is removed).} 
\label{fig:graph_comparison} 
\end{figure*}

As shown in Table \ref{cumulative_loc} and Table \ref{ram_usage}, re-localization performance for hand-crafted features with graph reduction (1-2-3-4-5-6 map) is better with less nodes than on single and 1+6 maps, but is lower than on 1+3+5, 2+4+6 and 1+2+3+4+5+6 maps. However, there is significantly less memory used when graph reduction is enabled. 
Another observation is that the average re-localization jumps are higher on 1-2-3-4-5-6 than with other multi-session maps. A first reason is that with graph reduction, the number of visual inliers is lower (at similar level than with single maps) because there are less frames that would have exactly the same illumination level than the frame to re-localize. Another reason is that maps with graph reduction would be less correctly optimized (i.e., not representing as well the environment than other multi-session maps), as there are less constraints in the graph. Without graph reduction, more odometry links are kept in the graph (VIO generates more accurate transforms between frames than re-localization using only RGB-D data), thus the map would be better optimized. To test this hypothesis, as ground truth is not available for this dataset, the map 1+2+3+4+5+6 has been reprocessed offline to add more links between all sessions. For each node in the graph, the closest node not already linked to it is tested with the TE approach. If TE is accepted, a new loop closure is added to graph. This whole process is repeated five times with all nodes in the map. The resulting map is then expected to be even closer to a real ground truth because of the added constraints. To make sure of this, the generated dense point cloud is inspected qualitatively to validate that there are no double surfaces or objects. Table \ref{ate_performance} shows the absolute trajectory error (ATE) \citep{sturm2012benchmark} results with and without graph reduction. The ATE is smaller on the maps without graph reduction because all constraints of all sessions are kept. Figure \ref{fig:graph_comparison} illustrates the error by superposing the optimized poses (blue nodes) on the ground truth poses (gray nodes). With graph reduction, the blue and corresponding gray nodes are less overlapping, meaning that the final optimized graph represents less well the environment, thus higher re-localization jumps would be expected, as observed in Table \ref{cumulative_loc}.

\begin{table}[!t]
\caption{ATE (mm) comparison with and without graph reduction}
\label{ate_performance}
\centering
\begin{tabular}{lcccccccc}
Map & SU & SI & BF & BK & KA & DY & SP & \blue{SG}\\
\hline
1+2+3+4+5+6 & 17 &  21 &   17 &  20  &  25 & 19 &  13 & 13 \\
1-2-3-4-5-6 & 41 &  63 &   68 &  64 &  48 & 57&  49 & 56 \\
\hline
\end{tabular}
\end{table}

\subsection{Consecutive Session Re-Localization}

\begin{figure*}[!t] 
\centering 
\includegraphics[width=\textwidth]{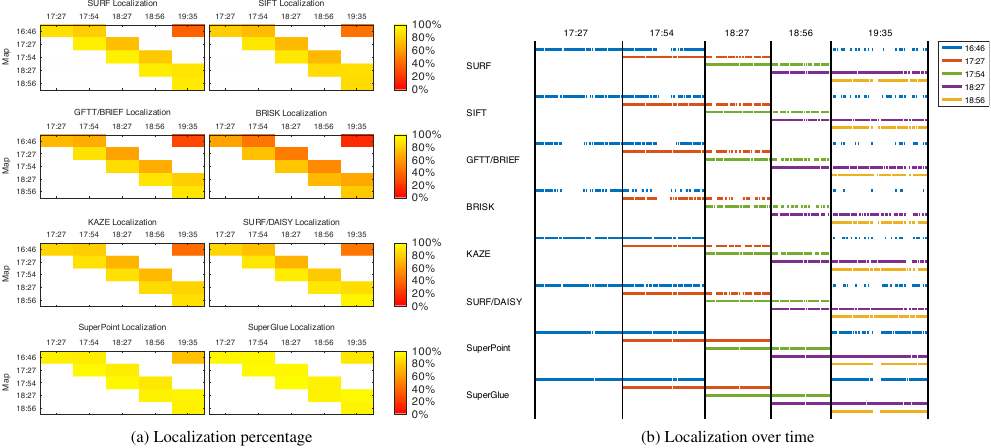} 
\caption{Re-lLocalization performance of the last five mapping sessions ($x$-axis) over preceding mapping sessions ($y$-axis), in relation to visual features used: a) re-localization percentage; b) re-localization over time.} 
\label{fig:consecutive} 
\end{figure*}

Results presented in Section \ref{sec:multisession} suggest that the best re-localization results are when using the six mapping sessions merged together. Having six maps to record before doing navigation can be a tedious task if an operator has to teleoperate the robot many times and at the right time. It would be better to ``teach'' once the trajectory to follow and have the robot repeat the process autonomously for the mapping sessions. The problem is if the robot cannot re-localize robustly on its previous trajectory, it may not be able to reproduce it completely, thus failing at capturing the required data. Figure \ref{fig:consecutive} shows re-localization performance using a previous mapping session. The diagonal values represents the case when localization occurs every 30 minutes using the previous map. Results just over the main diagonal are if localization is done each hour using a map taken one hour ago (e.g., for the 1+3+5 and 2+4+6 multi-session cases). The top-right lines are for the 1+6 multi-session case during which the robot would be activated only at night while trying to re-localize using the map learned during the day. Having low re-localization performance is not that bad, but re-localizations should be evenly distributed otherwise the robot may get lost before being able to re-localize after having to navigate using dead-reckoning over a small distance. The maximum distance that the robot can robustly recover from depends on the odometry drift: if high, frequent re-localizations would be required to correctly follow the planned path. Looking at \figurename \ref{fig:consecutive}b, SURF, SIFT, KAZE, DAISY and SuperPoint \blue{(with or without SuperGlue)} are features that do not give large gaps if maps are taken 30 minutes after the other. For maps taken 1 hour after the other, only KAZE, SuperPoint and DAISY do not show large gaps. Finally, SuperPoint may be the only one that could be used to only map the environment twice (e.g., one at day and one at night) and re-localize robustly using the first map. \blue{Table \ref{gap_distance} shows the largest distance (gap) in meters that the robot would have travelled on dead-reckoning in \figurename \ref{fig:consecutive}b, depending if the maps were taken 30 min, 60 min or 120 min apart. The percentage shows how many frames were re-localized under 55 cm of the previous re-localization. The number 55 has been chosen as the maximum distance between two consecutive frames taken at 1 Hz while walking at 55 cm per second.}

\begin{table}[!t]
\caption{\blue{Maximum distance (m) travelled while not being re-localized and the percentage of frames re-localized under 55 cm of the last re-localization}}
\label{gap_distance}
\centering
\begin{tabular}{lrrrrrrrr}
Interval between maps & SU & SI & BF & BK & KA & DY & SP & SG\\
\hline
\multirow{2}{*}{30 min} & 1.45 &  1.45 &   2.94 &  1.92  &  1.72 & 1.58 &  0.98 & 0.84 \\
       & 93\% &  89\% &   90\% &  92\%  &  90\% & 92\% &  95\% & 97\% \\
\hdashline[0.5pt/5pt]
\multirow{2}{*}{60 min}  & 4.50 &  3.59 &   4.68 &  4.68 &  4.65 & 4.65 &  1.38 & 1.11 \\
       & 78\% &  74\% &   73\% &  68\%  &  74\% & 74\% &  90\% & 95\% \\
\hdashline[0.5pt/5pt]
\multirow{2}{*}{120 min}  & 4.50 &  4.75 &   5.82 &  5.82 &  4.79 & 4.65 &  2.74 & 1.11 \\
       & 29\% &  34\% &   28\% &  20\%  &  38\% & 34\% &  72\% & 87\% \\
\hline
\end{tabular}
\end{table}

\section{DISCUSSION}
\label{sec:discussion}

Multi-session seems a valid approach to improve visual re-localization robustness to illumination changes in indoor environments. The dataset used in this paper is however limited to one day. Depending whether it is sunny, cloudy or rainy, or because of variations of artificial lighting conditions in the environment or if curtains are open or closed, more mapping sessions would have to be taken to keep high re-localization performance over time. During weeks or months, changes in the environment (e.g., furniture changes, items that are being moved, removed or added) could also influence performance.
Continuously updating the multi-session map to adapt over time to environment changes could be a solution \citep{labbe2017} which however, as the results suggest, would require more RAM even if graph reduction is enabled. For very long-term continuous multi-session mapping, a solution using RTAB-Map could be to enable its memory management approach \citep{labbe13appearance}, which would limit the size of the map in RAM. Another complementary approach to graph reduction could be to remove offline nodes on which the robot did not re-localize for a while (like weeks or months). Each node in the map would have to keep information about when was the last time a new frame has been re-localized on it. For example, if a room in the house has been renovated or redecorated, the robot could eventually definitely ``forget'' the old room images while keeping only the new ones. \blue{Similarly, the more formal probabilistic approach to model feature persistence from \citep{rosen2016towards} could also be integrated to remove features from the map that have ``vanished" over time from the environment.} 

To construct the multi-session map from consecutive sessions, we almost followed exactly the same trajectory every time, so the camera orientation and position were very similar between the trajectories. On a robot, it may be not always the case. If the robot has to avoid a dynamic obstacle and moves out of its trajectory, even if the odometry is accurate, it may get lost because the point of view of the robot would be too different from the ones in the map (assuming the robot has only a single camera with limited field of view). After detecting that the robot cannot plan for a while in the map (because it has drifted too much), a way to recover from this could be to plan a path in the center of the current room, independently of the global map, by using only the local map around the robot. This would work only if the center of the rooms has been captured in the mapping sessions. To do so, during the initialization of the first session of the multi-session map, the robot should follow general navigation rules like staying as far as possible of obstacles, so naturally it would map center of corridors and rooms, which will be easier afterwards to re-localize on when appending new sessions with different illumination levels. \blue{Instead having the robot re-mapping multiple times the environment, a digital twin of the target environment could be created to simulate illumination variations. In \citep{caselitz2020camera}, all possible lighting variations (based on combinations of lamps that can be on or off) including shadows are simulated in real-time using the latest ray tracing technology. The camera can then be robustly tracked in the environment even if lights are turned off or on (creating drastic changes of illumination) during re-localization. Re-localization could then go beyond the recorded trajectory with same points of view. However, if the environment structurally changes, the digital twin would need to be updated at some point, which may not be as simple as recording a new mapping session with the robot itself.} 

In terms of limitations, this visual re-localization approach would obviously not work in perfect darkness. \blue{RTAB-Map can use LiDAR or ToF (Time of Flight) camera geometric data to refine re-localization's transformation estimation \citep{labbe18jfr}, but it cannot do global re-localization without the discriminative visual features of a standard camera.} \blue{This visual-based approach} could be compatible with a camera system or robot equipped with lights, but it has yet to be tested. On some applications, the re-localization jump errors (around 2-6 cm) presented in the results may be also too high. As observed, the higher the number of inliers in TE is, the lower the re-localization jumps would be (greater accuracy). The Vis/Inliers parameter could be increased to accept only re-localization with higher number of inliers, at the cost of less frames re-localized. \blue{Note that re-localizing less often (creating large gaps of dead-reckoning) will produce higher re-localization jumps and also increase the chance to become completely lost. There is then a trade-off to think about.}

\blue{\citep{bai2019survey} suggest that neural networks in visual SLAM are becoming as competitive and even better than classical approaches. While end-to-end localization approach like PoseNet \citep{kendall2015posenet} is not currently as competitive as classical SLAM approaches in indoor setting, replacing parts of the classic pipeline by their neural network counterparts can indeed increase robustness. Results in our paper suggest that the usage of SuperPoint as feature detector increases the overall performance of re-localization against illumination changes. As mentioned in Section \ref{sec:related_work}, the usage of a learned global descriptor (like NetVLAD \citep{arandjelovic2016netvlad}) could also improve likelihood accuracy in replacement of BOW. The integration in this paper of SuperGlue for feature matching helps to get more feature correspondences than the classic nearest neighbor approach when illumination is different between mapping and localization sessions. For transformation estimation, an approach such as DSAC \citep{brachmann2021visual} could be used to improve re-localization accuracy in replacement of the classic PnP RANSAC approach used in this paper. While the robustness to illumination is significantly increased using neural networks, computational requirements exposed in this paper show that they may not be used efficiently on all systems. If the system capabilities are limited (e.g., no GPU), RTAB-Map could still rely on classic methods using the proposed multi-session approach to get similar robustness to illumination, at the cost of having more sessions to capture. However, if the system can run those neural networks, it is recommended to use them with RTAB-Map to decrease the number of recorded sessions required for optimal re-localization performance.}

\section{CONCLUSION}
\label{sec:conclusion}

Results in this paper suggest that regardless of the visual features used, similar re-localization performance is possible using a multi-session approach. The choice of the visual features could then be based on computation and memory cost, specific hardware requirements (like a GPU) or licensing conditions. The more illumination invariant the visual features are, the less sessions are required to reach the same level of performance. Graph reduction can further decrease significantly  memory usage of multi-session maps while keeping high re-localization performance, at cost of slightly worst re-localization accuracy. As an improvement, a better selection of which nodes to keep in the multi-session map using a strategy described in \blue{\citep{muhlfellner2016summary, halodova2019predictive}} may help improve re-localization performance and accuracy when graph reduction is applied. 

In future works, we plan to test this approach on a real robot to study if multiple consecutive sessions could indeed be robustly recorded autonomously with standard navigation algorithms. Testing over multiple days and weeks could give also a better idea of the approach's robustness on a real autonomous robot. The outdoor RobotCar \citep{RobotCarDatasetIJRR} \blue{or NCTL \citep{carlevaris2016university}} datasets could be used to evaluate if the same conclusions can be applied to outdoor scenarios including seasonal changes.


\section*{Funding}
This work is partly supported by the Natural Sciences and Engineering Research Council of Canada.

\bibliographystyle{apalike}
\bibliography{Manuscript}

\end{document}